\documentclass[conference]{IEEEtran}

\IEEEoverridecommandlockouts

\usepackage{times}
\usepackage{epsfig}
\usepackage{graphicx}
\usepackage{amsmath}
\usepackage{amssymb}
\usepackage[nolist]{acronym}
\usepackage{xcolor}
\usepackage{cite}
\usepackage{mathtools}  
\usepackage{tabulary}
\usepackage{lipsum}
\usepackage{bbm} 
\usepackage{comment} 

\usepackage[hidelinks]{hyperref}
\definecolor{darkred}{RGB}{150,0,0}
\definecolor{darkgreen}{RGB}{0,150,0}
\definecolor{darkblue}{RGB}{0,0,150}
\hypersetup{colorlinks=true, linkcolor=red, citecolor=blue, urlcolor=darkblue}

\begin{document}

\title{
Controlling the Latent Space of GANs through Reinforcement Learning: A Case Study on Task-based Image-to-Image Translation
}

\author{  Mahyar Abbasian$^{*1}$, Taha Rajabzadeh$^{*2}$, Ahmadreza Moradipari$^3$, Seyed Amir Hossein Aqajari$^1$, \\ Hongsheng Lu$^3$, and Amir M. Rahmani$^1$
%
\thanks{$^*$Authors have equal contribution.
$^1$University of California, Irvine, {\tt\small \{abbasiam,  saqajari, a.rahmani\}@uci.edu}.
$^2$Stanford University, {\tt\small tahar@stanford.edu}.
$^3$Toyota Motor North America, InfoTech Labs, {\tt\small \{Hongsheng.lu, ahmadreza.moradipari\}@toyota.com}.}
}

\maketitle
\begin{abstract}
Generative Adversarial Networks (GAN) have emerged as a formidable AI tool to generate realistic outputs based on training datasets. However, the challenge of exerting control over the generation process of GANs remains a significant hurdle.
In this paper, we propose a novel methodology to address this issue by integrating a reinforcement learning (RL) agent with a latent-space GAN (l-GAN), thereby facilitating the generation of desired outputs. More specifically, we have developed an actor-critic RL agent with a meticulously designed reward policy, enabling it to acquire proficiency in navigating the latent space of the l-GAN and generating outputs based on specified tasks.
To substantiate the efficacy of our approach, we have conducted a series of experiments employing the MNIST dataset, including arithmetic addition as an illustrative task. The outcomes of these experiments serve to validate our methodology.
Our pioneering integration of an RL agent with a GAN model represents a novel advancement, holding great potential for enhancing generative networks in the future.  
\end{abstract}



\section{Introduction}
The Generative Adversarial Network (GAN) represents an unsupervised machine learning technique that exhibits the ability to generate synthetic data encompassing diverse domains, including but not limited to images \cite{StyleGAN}, videos \cite{nvidia}, and 1D bio-signals\cite{Harada_2019}. Recent progress in GAN research has yielded substantial advancements in the realm of generating high-quality, realistic images \cite{StyleGAN}.
The fundamental architecture of a GAN comprises a minimum of two deep neural networks engaged in a competitive relationship, operating within the confines of a shared loss function. These networks are commonly referred to as the Generator and the Discriminator. The Generator network synthesizes artificial images by leveraging random noise inputs that mimic the characteristics of genuine images. In contrast, the Discriminator network undertakes the task of discerning between authentic images from the dataset and synthetic counterparts generated by the Generator.
In contemporary times, GANs have found application in a myriad of real-world scenarios, particularly in image-to-image translation tasks, such as face attribute editing \cite{CycleGAN2017,DBLP:journals/corr/abs-1904-09709,DBLP:journals/corr/abs-1711-09020,DBLP:journals/corr/abs-1711-10678,DBLP:journals/corr/abs-1812-04948}, image reconstruction \cite{DBLP:journals/corr/abs-1904-09709, DBLP:journals/corr/abs-1801-07892}, image resolution manipulation \cite{DBLP:journals/corr/LedigTHCATTWS16,DBLP:journals/corr/abs-1908-06382}, and video streaming \cite{nvidia}. A primary challenge addressed by many of these solutions pertains to the interpretation of the latent space of the Generator. Consequently, the exploration of efficient methods to control and interpret the latent space of the Generator has garnered significant attention.

The existing body of literature pertaining to GANs' latent-space control and interpretation can be classified into three primary categories. The initial category encompasses approaches that involve the extraction of feature maps from images, followed by converting these feature maps into latent vectors. Subsequently, these latent vectors are employed to generate edited images. \cite{DBLP:journals/corr/abs-1904-09709, DBLP:journals/corr/abs-1907-11922, DBLP:journals/corr/abs-1811-10100, DBLP:journals/corr/abs-2007-05892, DBLP:journals/corr/abs-2011-11900}. Within this category, there is a predominant emphasis on face attribute editing, with considerable attention directed towards concurrently training the feature mapping alongside GAN training.
The second category centers around exploring the latent space to identify an appropriate interpretation for this space. This category further subdivides into supervised \cite{DBLP:journals/corr/abs-2102-12139} and unsupervised \cite{DBLP:journals/corr/abs-1711-10678, DBLP:journals/corr/ZhouXYFHH17, Yang_2021_CVPR} methods. In both categories above, namely the feature map extraction (category one) and latent space interpretation (category two), the training process occurs in tandem with the training of the Generator and Discriminator components of the GAN model. This simultaneous training enables the interconnected networks to optimize their respective objectives jointly, fostering a cohesive learning process.
The third category, closest to the present paper, explores the latent space to achieve desired outcomes. Specifically, this approach involves the initial training of the GAN model on a given dataset, after which it is treated as a black-box. Subsequently, a mechanism is devised to navigate the latent space \cite{DBLP:journals/corr/YehCLHD16} or employ techniques such as Reinforcement Learning (RL) to learn and optimize the latent space \cite{sarmad2019rl}. The primary objective of our research is to train an RL agent as a mechanism to acquire proficiency in learning and controlling the latent space of the GAN, thus enabling the execution of a specific set of desired tasks.  


The utilization of RL to govern the latent space of a GAN confers several advantages, particularly in the realm of multi-modal image-to-image translation. Notably, employing RL allows the RL agent to undergo training with respect to various reward policies, rendering it highly adaptable for diverse translation tasks.
Specifically, the RL agent receives an image as input and learns to effectively transform it into a desired parameter within the latent space of a pre-trained GAN, guided by a meticulously designed reward policy. Consequently, this process yields a translated image that aligns with the desired output.
This approach offers remarkable flexibility, as it facilitates seamless adaptation to new translation tasks by simply modifying the reward policy of the RL agent and subsequently retraining it rather than necessitating the retraining of the entire GAN architecture. 

Incorporating RL to govern the latent space of a GAN expedites the training process and mitigates the risks associated with convergence issues and instability that may arise during GAN retraining. In contrast, the conventional back-propagation method needs a dedicated learning process. It necessitates multiple iterations to ascertain each new image's desired latent space parameter, resulting in time-consuming and less accurate outcomes.
The application of RL in controlling GAN networks finds relevance in various domains, including mixed and augmented reality. One application involves style transfer, where GANs can generate real-time stylized images that overlay the real-world view with artistic styles. Another use is object generation, where GANs generate 3D virtual objects that can be placed and interacted with in mixed-reality environments. GANs can also synthesize realistic scenes, creating virtual scenes that match the real world and enhancing the mixed reality experience. Real-time virtual avatars can be generated using GANs, mimicking user facial expressions and movements for an immersive augmented reality experience. Additionally, GANs can generate environmental effects like weather conditions, allowing users to experience virtual changes in their surroundings. Overall, GANs offer versatile and powerful control over mixed and augmented reality applications, enabling enhanced user experiences and greater immersion.

To the best of our knowledge, the sole existing work exploring RL's utilization to control a GAN is presented in Sarmad et al. \cite{sarmad2019rl}. However, their proposed model is confined to the regeneration of 3D point clouds from corrupted data, thus needing more capacity for extension to more intricate domains, such as image-to-image translation applications. Image-to-image translation applications introduce distinctive challenges that warrant specific considerations.
Firstly, ensuring convergence in image-to-image translation necessitates imposing certain constraints on the model, which may involve incorporating additional parameters to extract pertinent image features effectively. These constraints are crucial for maintaining the stability and coherence of the translation process.
Moreover, defining and evaluating the policy in image-to-image translation applications entails additional complexities. Assessing the fidelity of the output image in relation to the target image becomes a nuanced task, requiring careful consideration of various metrics to gauge the degree of resemblance between the two.

In this paper, we propose an architecture that combines RL with a GAN and demonstrate it in a case study of image-to-image translation. Our proposed framework exhibits versatility, as it can be effectively applied to diverse applications, including image/video attribute editing and image/video compression.
Notably, we introduce a novel concept termed ``task-based image-to-image translation," which enables the generation of images based on specific tasks. This concept holds significant potential in dynamic image attribute editing scenarios, where distinct desired objectives can be specified as input. The RL component then governs the GAN model to generate the target image accordingly, adapting the translation process based on the defined goals.
Furthermore, the task-based image-to-image translation concept finds applicability in regulating video quality during streaming. By defining the task as the desired output quality resolution, the RL-controlled GAN can dynamically adjust the translation process to optimize video quality in real time.

To validate the effectiveness of our proposed architecture, we conduct experiments utilizing the widely used MNIST dataset. The training process involves training a GAN network on this dataset while incorporating an RL agent to learn the latent space of the GAN. This learning process is facilitated by utilizing an input image and the explicit designation of a task, which collectively guides the RL agent to generate an output image that conforms to the specified task requirements.
In our model, the task is specifically defined as the addition of a predetermined number to the input image. Consequently, the generated output image corresponds to the original input image augmented by the designated task number.

The contributions of this paper are as follows:
\begin{itemize}
\item Proposing a novel RL-controlled GAN for image-to-image translation.
\item Introducing the notion of task/objective through RL-controlled GAN.
\item Evaluating the performance of our proposed architecture using a publicly available dataset.
\end{itemize}

In the following sections, we provide a detailed description of our proposed method and its implementation. We then present the experimental results and discuss the performance of our model. Finally, we summarize our findings and potential avenues for future research in this area.

\section{Dataset}
This study employs the widely recognized MNIST handwritten digits dataset to assess the performance of our proposed architecture. The MNIST dataset encompasses a substantial collection of handwritten digits and is frequently employed in training image processing systems. Our deliberate choice of a simple dataset aims to showcase the effectiveness of our model in operating and mitigating risks associated with more intricate datasets. The training set of the MNIST dataset, comprising 60,000 samples, was divided into 80\% for training, 20\% for validation, and the remaining 10,000 samples were assigned to the test set.

\section{Method}
In this section, we explain the pipeline needed to train an RL agent to control the generation of the GAN. Each piece of the pipeline is explained in detail as follows:

\begin{figure}[t]
\includegraphics[width=9cm]{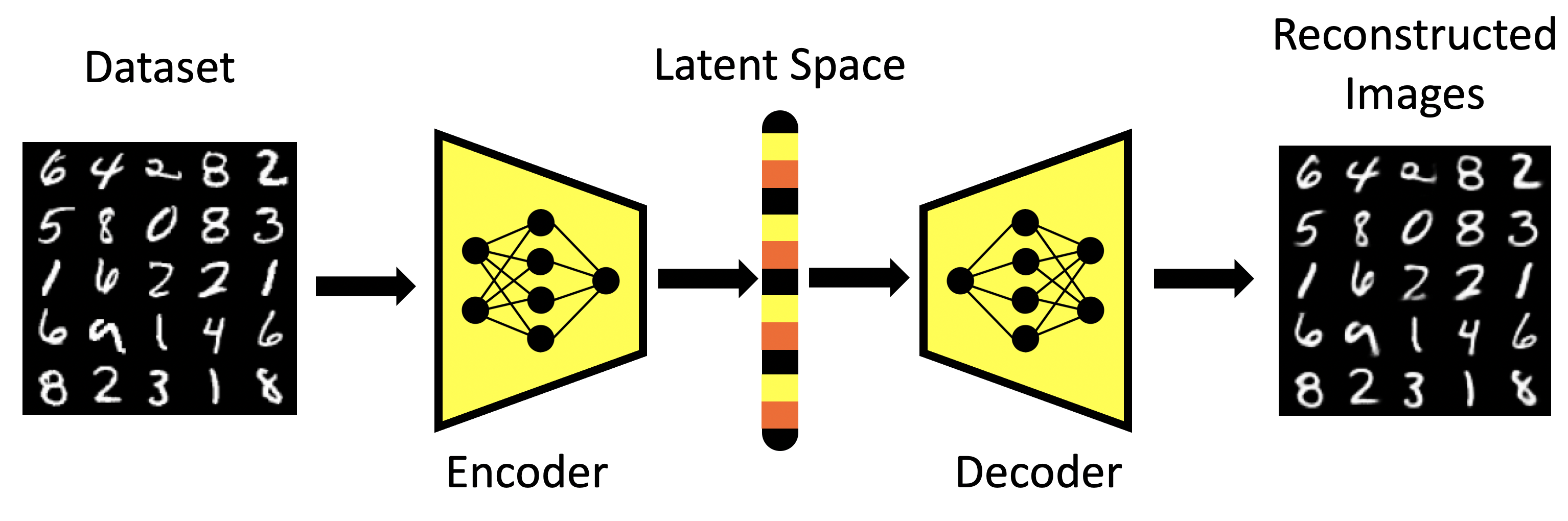}
\centering
\caption{The auto-encoder architecture\label{fig:ae_architecture}. Dataset images are encoded to latent space representation and are reconstructed back to the same image by the decoder.}
\end{figure}
\subsection{Auto-Encoder}
The Auto-Encoder model is utilized for encoding the dataset into a latent space, which facilitates the training of a stable GAN. This encoding process is also advantageous for training the RL agent, as it enables the agent to learn from a simplified representation of an image rather than the image itself. For the training of the auto-encoder, we normalized and directly employed the MNIST images. Our constructed auto-encoder, depicted in Figure \ref{fig:ae_architecture}, encodes 28 by 28 MNIST images into a latent space of size 32 through the $E$ transform and subsequently decodes them back to their original form using the $E^{-1}$ transform. The auto-encoder is trained using the Binary Cross-Entropy (BCE) loss as the reconstruction loss.
\begin{equation}
    \begin{split}
        loss_{AE} = -\text{Mean}\{E^{-1}(E(x))\log{x} \\
        + (1-E^{-1}(E(x)))\log{(1-x)} \}
    \end{split}
\end{equation}
Following the training of the Auto-Encoder, we detach the encoder to encode and transform the dataset into its latent space representation, which aids in the training of both the GAN and the RL agent. Moreover, the decoder visualizes the encoded data, such as the generator output.
\subsection{$l$-GAN}
Inspired by \cite{sarmad2019rl}, a latent-space GAN (l-GAN) framework is devised to generate latent representations of data derived from the MNIST dataset. In our approach, we utilized the pre-trained Encoder of the Auto-Encoder to encode MNIST images into a 32-dimensional vector. This preprocessing step aims to reduce the computational burden on the network and simplify the training process, thereby achieving faster and more efficient training. Consequently, the Generator and Discriminator are trained using the dataset encoded with the Auto-Encoder from the previous section.

The design of the Generator and Discriminator models follows \cite{zhang2019self}, which utilizes the Self-Attention Generative Adversarial Network (SAGAN) structure. The critical factor in this design is incorporating query, key, and value layers to generate attention coefficients for each element in the generator. This attention mechanism, as proposed in \cite{zhang2019self}, enables the model to generate results that closely resemble real data by focusing on important parts of the input. Essentially, the data is viewed as a key-value pair, and the output is obtained as a weighted sum of the values based on the dot product of the keys and a query layer obtained from the previously hidden layer $x$, represented as $\boldsymbol{k(x) = W_k x}$ and $\boldsymbol{q(x) = W_q x}$, respectively.
The weights are calculated as follows: 
\begin{equation*}
\small
\beta_{j, i}=\frac{\exp \left(s_{i j}\right)}{\sum_{i=1}^{N} \exp \left(s_{i j}\right)}, \text { where } s_{i j}=\boldsymbol{k}\left(\boldsymbol{x}_{i}\right)^{T} \boldsymbol{q}\left(\boldsymbol{x}_{\boldsymbol{j}}\right)
\end{equation*}
With the weight output layer is given from value layer $\boldsymbol{v}$ as:
\begin{equation*}
\small
\boldsymbol{o}_{j}=\boldsymbol{v}\left(\sum_{i=1}^{N} \beta_{j, i} \boldsymbol{h}\left(\boldsymbol{x}_{\boldsymbol{i}}\right)\right), \boldsymbol{h}\left(\boldsymbol{x}_{\boldsymbol{i}}\right)=\boldsymbol{W}_{\boldsymbol{h}} \boldsymbol{x}_{\boldsymbol{i}}, \boldsymbol{v}\left(\boldsymbol{x}_{\boldsymbol{i}}\right)=\boldsymbol{W}_{\boldsymbol{v}} \boldsymbol{x}_{\boldsymbol{i}}
\end{equation*}
The loss for training models is a hinge loss and is calculated as follows:
\begin{equation*}
\small
\begin{aligned}
L_{D}=&-\mathbb{E}_{x \sim p_{\text {data }}}[\min (0,-1+D(E(x)))] \\
&-\mathbb{E}_{z \sim p_{z}}[\min (0,-1-D(G(z)))] \\
L_{G}=&-\mathbb{E}_{z \sim p_{z}} D(G(z))
\end{aligned}
\end{equation*}
The notation $E(\cdot)$ represents the Encoder of the pre-trained Auto-Encoder model mentioned in the previous section, while $G(\cdot)$ and $D(\cdot)$ correspond to the Generator and Discriminator, respectively.

Another notable aspect of this design is the application of spectral normalization to both the Generator and Discriminator. Spectral normalization, as advocated by \cite{zhang2019self}, is employed to enhance the stability of the model by constraining the magnitudes of the weights. This normalization technique prevents the values from becoming excessively large and mitigates the occurrence of abnormal gradients.

\begin{figure}[t]
\includegraphics[width=0.5\textwidth]{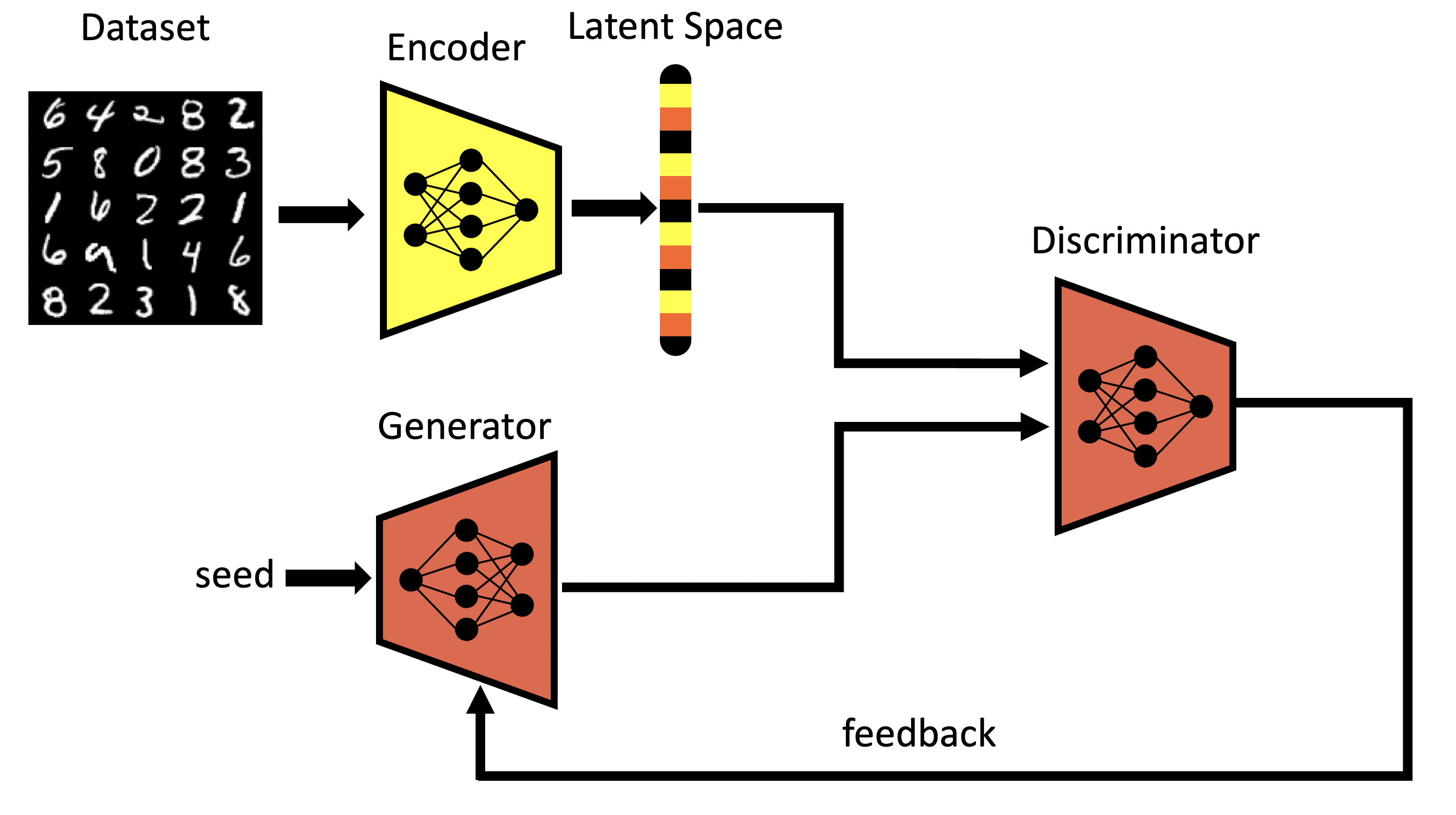}
\centering
\caption{The $l$-GAN architecture\label{fig:gan_architecture}.}
\end{figure}
\subsection{RL Agent}
The RL design in this study follows the Twin Delayed Deep Deterministic Policy Gradients (TD3) model introduced by \cite{fujimoto2018addressing}. The TD3 model is essentially an Actor-Critic policy gradient method comprising two distinct components. Firstly, the Critic, which adopts a value-based approach, aims to estimate the value function based on the state and action. Secondly, the Actor, which employs a policy-based approach, seeks to determine the optimal policy from the state directly. The Actor-Critic framework combines these two methods, where the Actor generates actions based on the state, and the Critic provides Q-values for those actions. When employing temporal difference learning, the Actor also utilizes the Critic's output to estimate the value of the next episode for updating purposes.

However, conventional value-based methods like Q-learning encounter an issue of value overestimation, attributed to the greedy maximum approach used to estimate Q-values based on available samples. This approach exhibits high variance and tends to produce larger values than expected. In addressing this problem, 
\cite{fujimoto2018addressing} propose a technique similar to double learning, which mitigates overestimation by incorporating two separate Q-value estimations and selecting the maximum index from one estimation and its corresponding value from the other estimation. The minimum value between these two separate estimations is chosen to avoid overestimation.

Additionally, this method introduces noise to the Q-values to encourage and regularize a smoother policy. The loss function for updating the critic, denoted as $\theta_i$, for each estimation ($i\in{1,2}$), is defined as:
\begin{equation*}
\small
L_c=N^{-1} \sum\left(y-Q_{\theta_{i}}(s, a)\right)^{2},
\end{equation*}
where $s,a$ are state and action and $y$ is estimated target Q value that is calculated as:
\begin{equation*}
\small
\begin{aligned}
y &=r+\gamma \min _{i=1,2} Q_{\theta_{i}^{\prime}}\left(s^{\prime}, \pi_{\phi^{\prime}}\left(s^{\prime}\right)+\epsilon\right), \\
\epsilon & \sim \operatorname{clip}(\mathcal{N}(0, \sigma),-c, c),
\end{aligned}
\end{equation*}
where $r$ is reward for noisy action and $\theta^{\prime}_i$ are target Q value estimations that are updated from $\theta_i$ by a factor $\tau$ as:
\begin{equation*}
\small
\theta^{\prime}_i = \tau \theta^{\prime}_i +(1-\tau)\theta_i
\end{equation*}
The Actor model is updated according to:
\begin{equation*}
\small
\nabla_{\phi} J(\phi)=\left.N^{-1} \sum \nabla_{a} Q_{\theta_{1}}(s, a)\right|_{a=\pi_{\phi}(s)} \nabla_{\phi} \pi_{\phi}(s)
\end{equation*}
Where $\pi_{\phi}(s)$ is policy (Actor model) and $ Q_{\theta_{1}}$ is estimation of Q value from Critic . The target network is updated by:
\begin{equation*}
\phi^{\prime} \leftarrow \tau \phi+(1-\tau) \phi^{\prime}
\end{equation*}

\subsection{Environment}
\begin{figure}[t]
\includegraphics[width=0.5\textwidth]{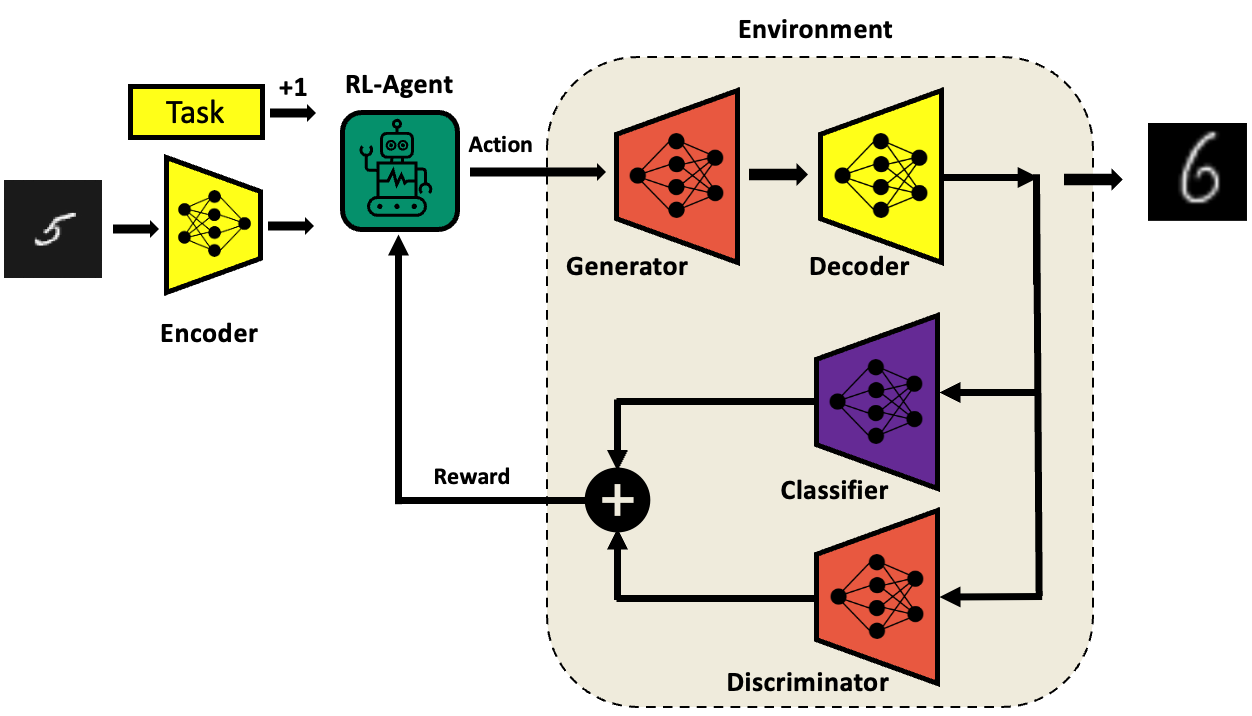}
\centering
\caption{ The environment is built by the generator, decoder, classifier, and Discriminator. All these components are pretrained on the dataset. Then, the RL agent takes the encoded image and task as input and selects a parameter from the latent space of the generator to generate the desired image. The classifier and Discriminator measure the correctness and realness of the generated image and return a reward to the RL agent.\label{fig:RL_architecture}}
\end{figure}

To facilitate the training of the RL agent, we construct an environment as depicted in Figure \ref{fig:RL_architecture}. During each episode, the RL agent receives an encoded image and a randomly generated number between 0 and 9, referred to as the task. Subsequently, the agent selects an action that shares the exact dimensions as the Generator input, to generate an encoded image that combines the number from the input image and the task. Consequently, the Decoder then reconstructs the encoded generated image, passed to both the Classifier and the Discriminator.
On the one hand, the Classifier verifies that the generated image corresponds to the number expected from a given task. On the other hand, the Discriminator evaluates the authenticity of the generated image. The weighted sum of the Classifier and Discriminator outputs is returned to the RL agent as the reward within the environment.

\section{Experiments}
In order to assess the control capabilities of the RL agent over the GAN's generation process, we conduct a series of experiments to train each component of the AI model. These experiments aimed to demonstrate that the RL agent can receive an input image of a specific number and generate a new image corresponding to the number resulting in the arithmetic addition of the input number and the task. For instance, when presented with an image of the number 5 and a task of +2, the RL agent should be able to identify a suitable seed for the generator to produce an image of the number 7.
The training process was executed on a single Quadro RTX 5000 GPU with 16GB of memory, utilizing 2 CPUs and approximately 5GB of RAM.
\subsection{Auto-Encoder}
We conducted the training of the Auto-Encoder using the normalized MNIST training dataset. The Encoder takes in 28 by 28 MNIST images and encodes them into a 32-length array, representing the latent space representation. The Auto-Encoder is trained using the Binary Cross-Entropy (BCE) loss, employing a batch size of 1024 and a learning rate of 0.002 for 20 epochs.

\begin{figure}[t]
\includegraphics[width=0.45\textwidth]{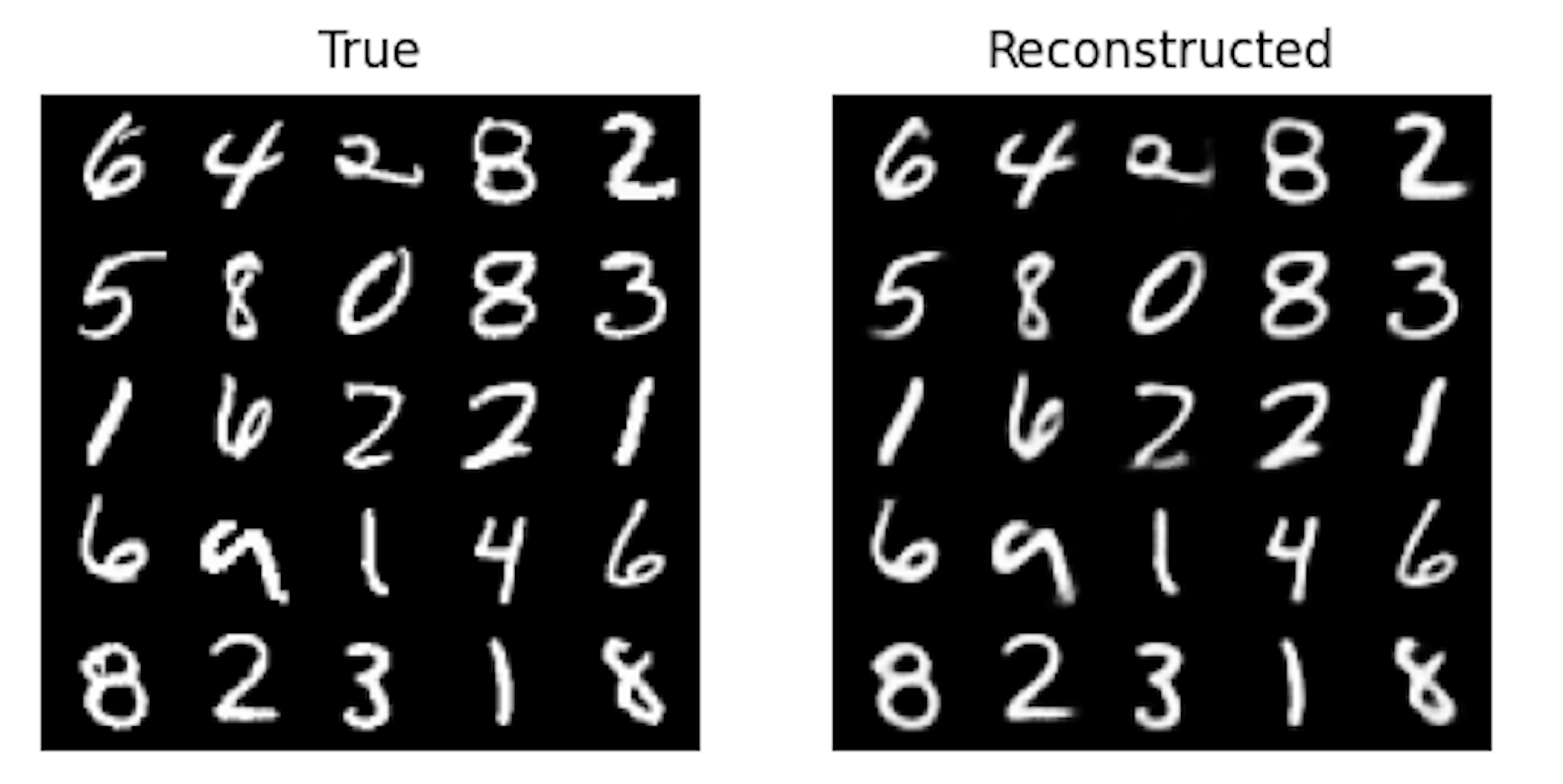}
\centering
\caption{The auto-encoder test result. The actual images are the input of the auto-encoder, successfully reconstructed by the auto-encoder. \label{fig:ae_result}}
\end{figure}

In Figure \ref{fig:ae_result}, we present the visualization of the Auto-Encoder's input and output. Notably, the Auto-Encoder effectively reconstructs the input images, indicating that the Encoder is prepared for utilization within the GAN and RL architectures for encoding input images.

\subsection{$l$-GAN}
The GAN model is trained using a latent space representation of the dataset derived from MNIST. As shown in Figure \ref{fig:gan_architecture}, this is achieved by incorporating the pre-trained Encoder as a transformation function applied to the dataset, creating a 32-length array representing the input data. We also consider a seed set as a generator's input to create random noise, and we note the generator's input as a GAN's latent space denoted by a vector $z$. Initially, the input $z$ (Generator input) dimension was set to 2. However, upon evaluating the results of the RL experiments, it became apparent that the trained model lacked diversity. Consequently, the dimension of $z$ was modified to 5 to enhance the variation of generated samples.
As described in the methodology, the hinge loss is utilized for training both the Generator and Discriminator.

\begin{figure}[t]
\includegraphics[width=0.45\textwidth]{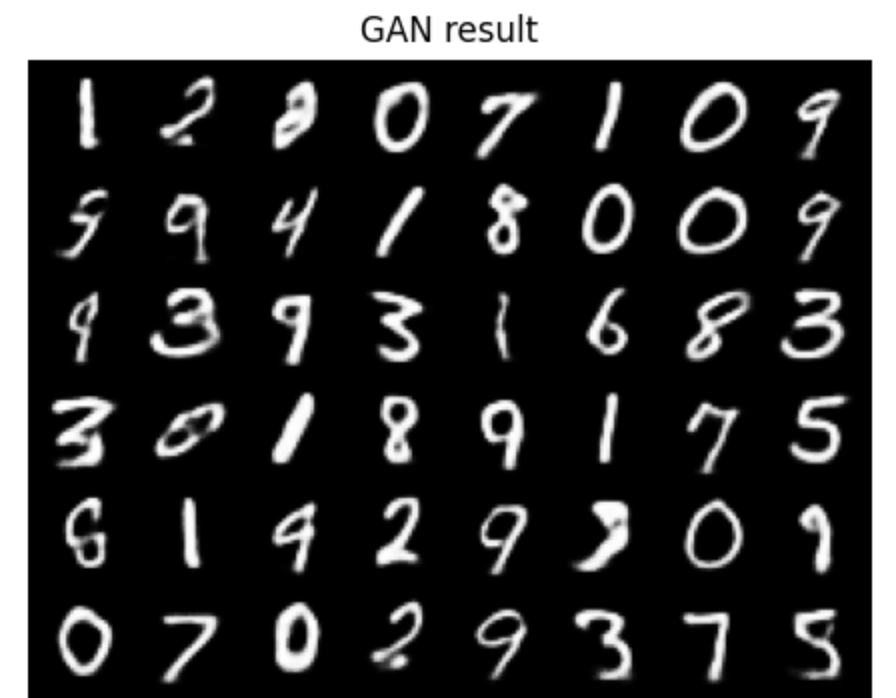}
\centering
\caption{GAN sample results. 48 images that are generated by generator.\label{fig:gan_figure}}
\end{figure}

A batch size of 50 was selected during training, and the Generator and Discriminator models were trained for 500,000 steps. The learning rate for both models was set to 0.00005. The hidden layer channel number was set to 16 for both the Generator and Discriminator.
Several results are visualized in Figure \ref{fig:gan_figure}.

\subsection{RL Agent}\label{SEC:RL_AGENT_experiment}
The final state of the RL model consists of a latent dataset similar to that of the GAN model. The action is represented by an array of length 5, with each dimension limited to a maximum absolute value of 1. The reward, as defined in the environment, is calculated based on the state's proximity, when passed through the Decoder, to the target value (which is the input label plus the given task). This proximity is measured by evaluating the results using the pre-trained Classifier's outputs. Additionally, the reward considers the accuracy of the resulting number, which is determined by the output of the pre-trained Discriminator.

During training, samples were drawn from the environment using a random policy for 50 steps. Subsequently, the policy was trained using a batch size of 50 for 500,000 steps, employing a learning rate 0.0005. Evaluation of the validation dataset was conducted every 10,000 steps.

\begin{figure}[t]
\includegraphics[width=0.48 \textwidth]{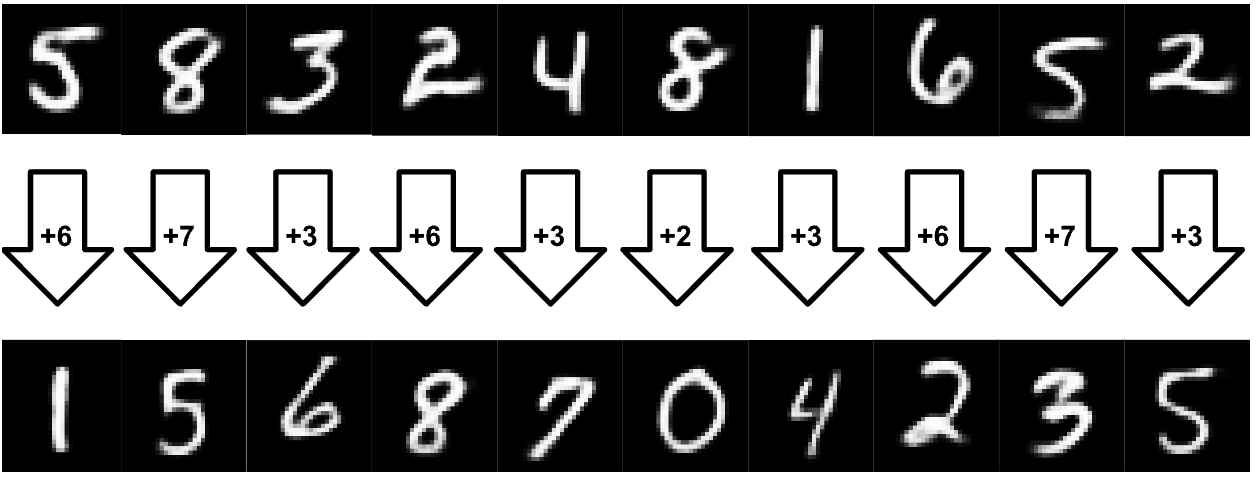}
\centering
\caption{RL sample results based on defined dynamic task. Each arrow shows the input task RL receives and converts the input image on the left to the output image based on the task. \label{fig:rl_results}}
\end{figure}

Figure \ref{fig:rl_results} showcases some results the trained RL agent achieved, demonstrating its success in producing the desired outcome (i.e., arithmetic addition of the task value to the input image). The pre-trained Classifier outputs the likelihood probability of its input image representing each of the ten numbers. As a result, as one of the likelihoods of the ten numbers approaches one (i.e., the likelihood measure), the Classifier is more confident that its input image corresponds to that particular number.

The reward policy is a weighted sum of the Classifier's output probability for the target number and the hinge loss of the Discriminator output and -1. As the Classifier's confidence increases, the probability of the target number approaches 1. Additionally, as the generated image becomes more realistic, the Discriminator's output tends to 0 from negative values. The reward formula is defined as follows:
\begin{equation}
 reward = \lambda_{cl} * \text{classifier$_i$} + \lambda_d * \text{discriminator} 
\label{eq:rl} 
\end{equation}
In equation \ref{eq:rl}, we consider the classifier weight, denoted as $\lambda_{cl}$=30, emphasizing the importance of the Classifier in the reward policy update. The target number, represented by $i$, corresponds to the number we expect the model to generate. The discriminator weight, denoted as $\lambda_d$=1, determines the significance of the Discriminator in the reward calculation. By assigning these weights, we aim to prioritize the role of the Classifier in assisting the RL agent in generating the desired target number.

\begin{figure*}[t]
\includegraphics[width=0.49\textwidth]{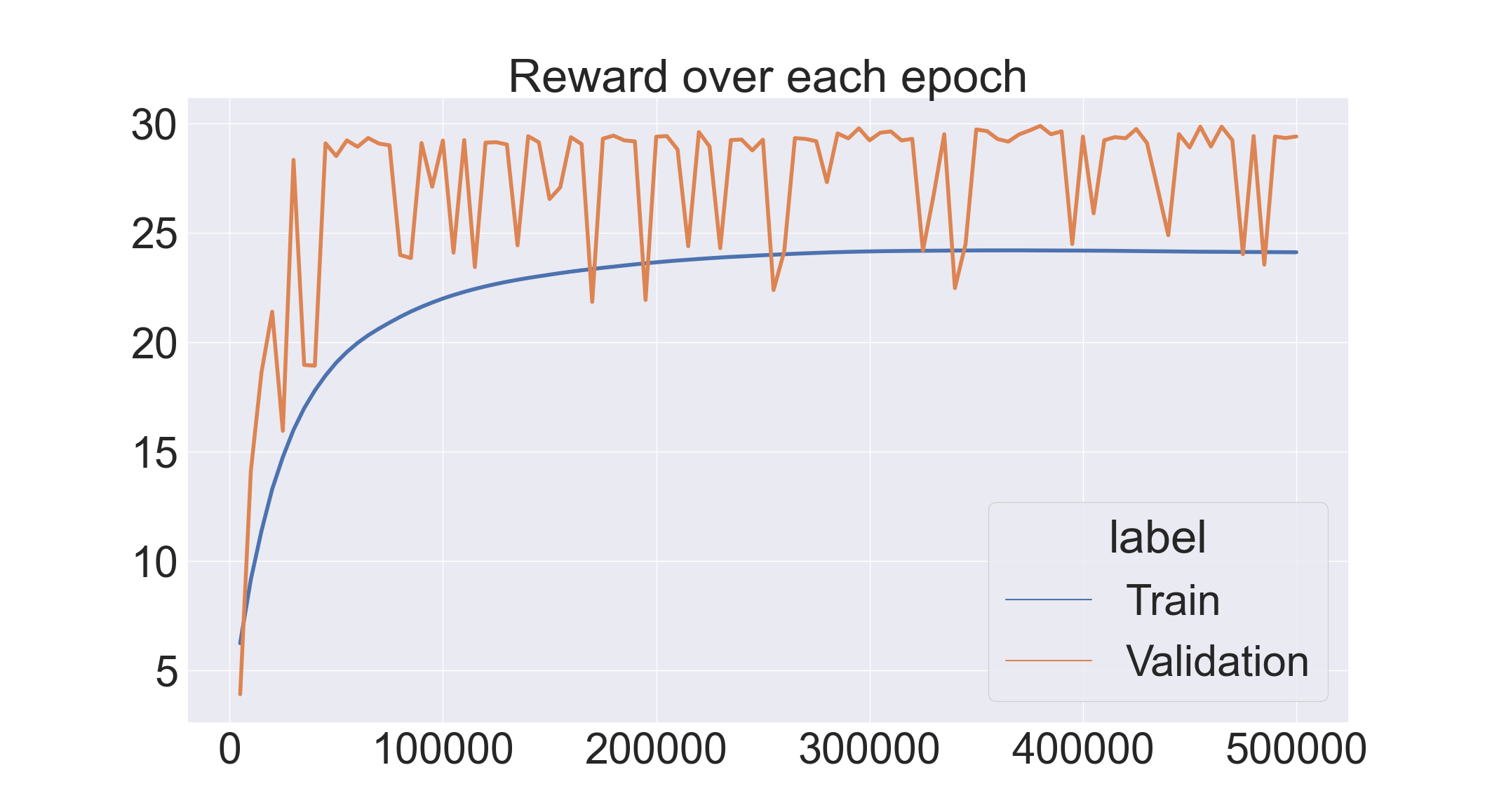}
\includegraphics[width=0.49\textwidth]{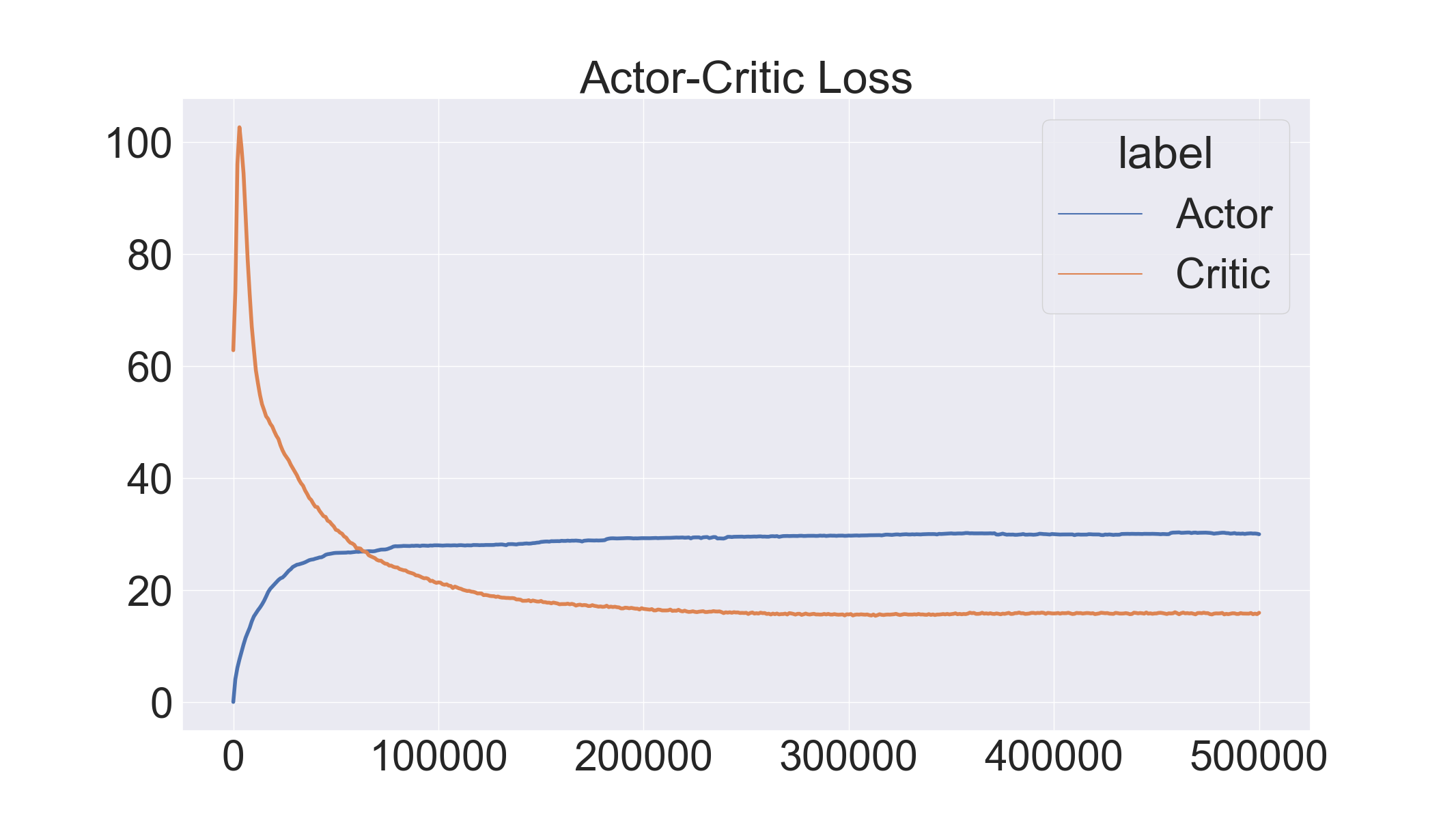}
\centering
\caption{Left: RL convergence to maximize the reward function. The blue line shows the training reward, and the red shows the validation after each epoch. Right: RL Actor-Critic loss convergence. The Actor is gradient ascending, and the Critic is gradient descending.}\label{fig:reward}
\end{figure*}

Figure \ref{fig:reward} (left) displays the training and validation reward values obtained during 500,000 epochs of model training. Based on our reward definition, the maximum reward is 30, and the RL agent exhibits convergence towards reaching this value. Notably, the validation reward demonstrates impressive results in most epochs, approaching a value close to 30.
Figure \ref{fig:reward} (right) depicts the Actor-Critic loss during the training process. The Actor loss follows a gradient ascent approach, while the Critic loss employs a gradient descent method. Both losses exhibit convergence over time.

\subsection{Evaluation}
In our model evaluation, we found that if we set the GAN's latent space dimension to 5, i.e., $z \in \mathbb{R}^5$ rather than 2, the model achieved a higher average overall reward. In fact, with a larger dimension, more features of the generated output can be captured, resulting in better generalization of the output domain and higher quality generation of images in general.

In Figure \ref{fig:reward} (left), the batch size used for the actor was 50, while for the overall training, it was set to 20 in our final run. The model exhibits convergence and accurately generates the next image based on the given task.
Figure \ref{fig:rl_results} presents a curated collection of sample inputs, corresponding tasks, and the generated images produced by our model, which shows the success of the trained RL agent. 

To assess the performance of our model, we consider a subset of the MNIST test set consisting of 10,000 images. For each test case, we selected an image from the dataset, randomly assigned a task (choosing a number between 0 and 9 to be added to the input image), and examined whether the generated image correctly reflected the desired result. The model achieved an impressive accuracy rate of 95.31\% in this evaluation.

To evaluate the robustness of the model against noise, we introduced Gaussian noise with a mean of 0 and a standard deviation of 0.3 to the input images. Under this noisy condition, the model achieved an accuracy rate of 81.79\%. It is worth noting that the decrease in accuracy can be attributed to the fact that the noise introduced caused some of the numbers in the images challenging to decipher, even for human observers.
Table \ref{tab:rsesult} summarizes the final results obtained from our model.
\begin{table}[h]
\centering
\caption{Accuracy Results on MNIST Dataset}
\label{tab:rsesult}
\begin{tabular}{l | l | l}
\textbf{Method}  & \textbf{Number of Samples}  & \textbf{Accuracy}  \\ \hline
Normal Image          & 10000              & 95.31\%                        \\
Noisy Image  & 10000                & 81.79\%         \\ \hline          
\end{tabular}
\end{table}
We employed the Discriminator and Classifier outputs as performance metrics to determine the accuracy presented in Table \ref{tab:rsesult} and evaluate the image quality of the generated images from the test set. As explained in Section \ref{SEC:RL_AGENT_experiment}, we utilized the metric $\lambda_{cl} * classifier_i$ to assess the quality of the generated images. A higher value of the target image number approaching 30 indicates that the model is more confident in producing an image that closely matches the target. Thus, we can conclude that the quality of the generated images is reasonably high, as the classifier was able to classify them with a high degree of confidence.
Table \ref{tab:quality} presents the average classifier output for the images in the test set, which is 28.58. This output indicates that nearly all correctly generated images possess high quality, and the classifier expresses strong confidence in identifying the target number.

We leveraged the pre-trained Discriminator from the GAN model to further assess image quality. We fed both the test set images and the generated images through the Discriminator and compared the quality of their inputs and outputs. Table \ref{tab:quality} displays the average output values of the Discriminator for the test set images and the generated images. The averages for both sets are closely aligned, suggesting that the quality of the generated images is comparable to that of the given images. 
\begin{table}[h]
\centering
\caption{Generated Image Quality}
\label{tab:quality}
\begin{tabular}{l | l}
\textbf{Method}  & \textbf{Average}  \\ \hline
Classifier          & 28.58                              \\
Input Image Discriminator       & 0.71              \\      
Generated Image Discriminator       & 0.70                  \\ \hline          
\end{tabular}
\end{table}
\section{Conclusion}

In this work, we present a comprehensive and innovative approach to image generation by integrating Generative Adversarial Networks (GANs) and Reinforcement Learning (RL). Our proposed model harnesses the capabilities of GANs to produce realistic images while employing RL techniques to control the generation process and achieve desired outcomes precisely. We conducted experiments using the MNIST dataset to train and evaluate our model. The results showcased the model's effectiveness in applying tasks to input images and generating the corresponding outputs. We report the accuracy metrics and image quality evaluations to support our findings. The integration of RL with GANs opens up new possibilities for enhancing generative networks, with potential applications including real-time image or video quality control, privacy preservation through encoding and regeneration using RL in GAN's latent space, domain conversion by training GAN on one domain and RL on the other, and image attribute editing by incorporating specific task requirements. These future directions demonstrate the broad potential and versatility of our proposed model.

\bibliographystyle{IEEEtran}
\bibliography{egbib}

\end{document}